\documentclass[11pt,a4paper]{article}
\usepackage{authblk}
\usepackage[hyperref]{acl2019}
\usepackage{times}
\usepackage{latexsym}
\usepackage{csquotes}
\usepackage{amsmath}
\usepackage[11pt]{moresize}
\usepackage{anyfontsize}
\usepackage{setspace}

\usepackage{mdframed}

\usepackage{graphicx}
\usepackage{float}
\usepackage{tabu}
\usepackage{pifont}
\newcommand{\cmark}{\ding{51}}%
\newcommand{\xmark}{\ding{55}}%

\usepackage{url}

\aclfinalcopy 

\title{\textbf{RUN through the Streets: \\   A New Dataset and Baseline Models for Realistic Urban Navigation}}
\author[1]{\textbf{Tzuf Paz-Argaman}}
\author[1,2]{\textbf{Reut Tsarfaty}}
\affil[1]{Open University of Israel
}
\affil[2]{Allen Institute for Artificial Intelligence}
\affil[ ]{\tt \{tzufar,reut.tsarfaty\}@gmail.com}

\date{}

\begin{document}
  \maketitle
\begin{abstract}

Following navigation instructions in natural language  requires a composition of language, action, and knowledge of the environment. Knowledge of the environment may be provided via visual sensors or as a symbolic world representation referred to as a {\em map}. 
Here we introduce the {\em Realistic Urban Navigation} (RUN) task, aimed at interpreting  navigation instructions based on a real, dense, urban map. Using Amazon Mechanical Turk, we collected a dataset of 2515 instructions aligned with actual routes over three regions of Manhattan.
We  propose a strong baseline for the  task and empirically investigate which aspects of the neural architecture are important for the RUN success. Our results empirically show that {\em entity abstraction},  {\em attention over words and worlds}, and  a constantly updating {\em world-state}, significantly contribute to task accuracy.

\end{abstract}

\section{Introduction and Background}

The task of interpreting  and following natural language (NL) navigation instructions involves interleaving different  signals,  at the very least the linguistic utterance and the representation of the world. For example, in \enquote {turn right on the first intersection}, the instruction  needs to be interpreted, and  a specific object in the  world (the  intersection)   needs to be located in order to execute the instruction.
In  NL navigation studies, the representation of the world  may be provided  via visual sensors \cite{misra2018mapping,blukis2018following,nguyen2018vision,yan2018chalet,anderson2018vision} or as a symbolic world representation. This work focuses on navigation based on a symbolic world representation 
(referred to as a {\em map}).

Previous datasets for NL navigation based on a symbolic world representation,  HCRC \cite{anderson1991hcrc,vogel2010learning, levit2007interpretation} and SAIL \cite{macmahon2006walk,chen2011learning,kim2012unsupervised,kim2013adapting,artzi2013weakly,artzi2014learning,fried2017unified,andreas2015alignment}  present  relatively simple worlds, with a small fixed set of entities known to the navigator in advance. Such representations bypass the great complexity of real urban navigation,  which consists of long paths and an abundance of previously unseen entities of different types. 




In this work we introduce {\em Realistic Urban Navigation} (RUN), where we aim to interpret navigation instructions relative to a rich symbolic representation of the world, given by a real dense urban map. To address RUN, we designed and collected a new dataset based on OpenStreetMap, in which we align NL instructions  to their corresponding routes. 
Using Amazon Mechanical Turk, we collected  2515 instructions over 3 regions of Manhattan, all specified (and verified) by  (respective)  sets of humans workers.
This task raises several challenges. 
First of all, we assume a large world, providing long routes,  vulnerable to error propagation; secondly, we assume a rich environment, with entities of various different types, most of which are unseen during training  and are not known in advance;  finally,  we evaluate on the full route intended, rather than on last-position only.

We then propose a strong neural baseline for  RUN  
where we augment a standard encoder-decoder architecture with an entity abstraction layer, attention over words and worlds, and a constantly updating world-state.
Our experimental results and ablation study show that this architecture is indeed better-equipped to treat  {\em grounding}  in realistic urban settings than standard sequence-to-sequence architectures. 
%
Given this RUN  benchmark,  empirical results, and evaluation procedure, we hope to encourage further investigation into the topic of interpreting NL instructions in realistic and previously unseen urban domains.

       

 \begin{table*}[h]
 \scalebox{0.57}{
 \centering
\begin{tabular}{|l|llll|llllllll|}
\hline
                & \multicolumn{4}{c|}{\textbf{Symbolic World Representation}}                                                                                                                                                                                                                            & \multicolumn{8}{c|}{\textbf{NL Utterances}}                                                                                                                                                                                                                                                                                                                                                                                                                                                                                                                                      \\ \hline
\textbf{Map}    & \textbf{\begin{tabular}[c]{@{}l@{}}Tokens\\ in\\ Map\end{tabular}} & \textbf{\begin{tabular}[c]{@{}l@{}}Entities\\ in\\ Map\end{tabular}} & \textbf{\begin{tabular}[c]{@{}l@{}}Number \\ of\\ Tiles\end{tabular}} & \textbf{\begin{tabular}[c]{@{}l@{}}Size\\ ($km^{2}$)\end{tabular}} & \textbf{Paragraphs} & \textbf{Instructions} & \textbf{\begin{tabular}[c]{@{}l@{}}Unique \\ Tokens\\ in Paragraphs\end{tabular}} & \textbf{\begin{tabular}[c]{@{}l@{}}Avg. \\ Tokens per \\ Instruction\end{tabular}} & \textbf{\begin{tabular}[c]{@{}l@{}}Avg. \\ Verbs per \\ Instruction\end{tabular}} & \textbf{\begin{tabular}[c]{@{}l@{}}Avg. \\ Actions per \\ Instruction\end{tabular}} & \textbf{\begin{tabular}[c]{@{}l@{}}Instruction\\ with advmode\\ before ROOT\end{tabular}} & \textbf{\begin{tabular}[c]{@{}l@{}}Avg. Named\\ Entities per \\ Instruction\end{tabular}} \\ \hline
1               & 10,353                                                             & 1,612                                                                & 5,457                                                                 & 0.51                                                               & 159              & 874                   & 735                                                                            & 14.42                                                                              & 1.53                                                                              & 18.71                                                                               & 12.81\%                                                                                   & 1.93                                                                                      \\
2               & 9,829                                                              & 1,134                                                                & 4,935                                                                 & 0.46                                                               & 128              & 884                   & 727                                                                            & 12.46                                                                              & 1.31                                                                              & 12.81                                                                               & 9.16\%                                                                                    & 1.32                                                                                      \\
3               & 8,844                                                              & 1,051                                                                & 5,452                                                                 & 0.51                                                               & 102              & 757                   & 654                                                                            & 11.86                                                                              & 1.29                                                                              & 12.44                                                                               & 10.3\%                                                                                    & 1.44                                                                                      \\ \hline
\textbf{Corpus} & 29,026                                                             & 3,797                                                                & 15,844                                                                & 1.48                                                               & 389              & 2515                  & \textbf{1451}                                                                  & 12.96                                                                              & 1.38                                                                              & 13.71                                                                               & 10.78\%                                                                                   & 1.57                                                                                      \\ \hline
\end{tabular}
}
\caption{Data Statistics of RUN: statistics over different maps and the full corpus. The table is divided into features of the symbolic world representation and the written paragraphs.}
\label{tab:linguistic}
\end{table*}

\begin{table*}[ht]
\scalebox{0.758}{
\begin{tabular}{|l|l|l|}
\hline
\multicolumn{1}{|c|}{\textbf{Phenomenon}} & \multicolumn{1}{c|}{\textbf{instructions}} & \multicolumn{1}{c|}{\textbf{Example from RUN}}                                                                    \\ \hline
Reference to  unnamed entity              & 53.33\%                                    & Walk to the first \textbf{stoplight} and turn left heading south.                                          \\
Reference to  named entity                & 93.33\%                                    & Walk a little more and you will reach your destination on your right: \textbf{Fantastic Cafe}.             \\
Coreference                               & 10\%                                       & Pass the intersection and \textbf{it} will be the second building on your right.                           \\
Sequencing                                & 20\%                                       & Walk to the \textbf{next} intersection, turning right to Avenue B.                                         \\
Count                                     & 23.33\%                                    & Walk \textbf{5} buildings down the street, and you will see the mcdonalds.                                 \\
Egocentric spatial relation               & 26.67\%                                    & B\&H photo will be immediately \textbf{on your right} and that is where you want to be.                    \\
Imperative                                & 83.33\%                                    & \textbf{Go} through the intersection and  \textbf{follow} the road past the Kmart center. \\
Direction                                 & 66.67\%                                    & Make a \textbf{right} and go up one block to the light on West 33rd  street and 7th Avenue .               \\
Condition                        & 26.67\%                                    & we will continue walking \textbf{till} we  come to the east 7th street intersection.                       \\
Verification                        & 20\%                                       & On your left toward about the middle of the block \textbf{you'll see} Alphabet City.                        \\ \hline
\end{tabular}
}
\caption{Linguistic Analysis of RUN: we analyze 30 randomly sampled instructions in RUN. The table characterizes linguistic phenomena in RUN, categorized according to the catalogue presented in  \citet{chen2018touchdown}.}
\label{tab:touch}
\end{table*}

 \begin{table}[ht]
 
     \centering
     \scalebox{0.7}{
\begin{tabular}{l|cccc}
\hline\hline
     & \textbf{\#Entities} &  \textbf{\begin{tabular}[c]{@{}c@{}}\#Unique  \\ Entities \end{tabular}}   & \textbf{\#Tiles  }              & \textbf{\begin{tabular}[c]{@{}c@{}}Tiles Moved per\\ Sentence\textbackslash Paragraph \end{tabular}} \\ \hline
HCRC & *11.93             & *8.125               & *11.93 \footnotemark        & n/a\textbackslash *9.75                                                                               \\
SAIL & 22                & 0                           & 33.33                                    & 1.3\textbackslash  5.34                                                                               \\
TTW  & *62.6             & **0  & 25       & n/a\textbackslash 2.5                                                                                 \\
RUN  & 932               & 365                         & 1059.6  & 12.2\textbackslash 78.89                  \\                                          \hline\hline                 
\end{tabular}
     }
     \caption{Quantitative Comparison of  the HCRC \cite{anderson1991hcrc}, SAIL \cite{macmahon2006walk}, TTW \cite{de2018talk}, and the new  RUN Dataset. *We average over three randomly chosen maps. 
**{\citet{de2018talk} assume perfect perception:  all entities at each location are known in advance.}}
     \label{table:statistic_maps}
     
 \end{table}

%
%

\footnotetext[1]{The task defined by \citet{vogel2010learning} is of moving between entities only.}

\section{The  RUN  Task and Dataset}
\label{task}

In this work we address the task of following  a sequence of NL navigation instructions given  in colloquial language based on a dense  urban map. 
%

The {\em input} to the RUN task we define is as follows: 
(i)
    a map with rich details divided into tiles,
(ii)
    an explicit starting point, and
(iii)
    a sequence of navigation  instructions  we henceforth refer to as a navigation {\em paragraph }.
    We refer to each sentence as an {\em instruction}, and we assume that following the individual instructions  in the {\em paragraph} one by one will lead the agent to the intended end-point.
The {\em output} of RUN is the entire route described by the paragraph, i.e.,  all coordinates up to and including its end-point, pinned on the map.


In order to address  RUN  we designed and collected a novel dataset, henceforth the {\em RUN dataset}, which is based on \href{http://www.openstreetmap.org}{OpenStreetMap (OSM)}.\footnote{OSM is a free, editable, map of the whole world, that was built by volunteers, with millions of users constantly adding informative tags to the map.} The map contains rich layers and an abundance of entities of different types. 
Each entity is complex and can contain (at least) four labels: name, type, is\_building=y/n, and house number. An entity can spread over several tiles.
As the maps do not overlap, only very few entities are shared among them.
The RUN dataset aligns NL navigation instructions to coordinates of their corresponding route on the OSM map.

 \begin{figure*}[th]


  \centering
\scalebox{0.98}{
            \includegraphics[width=1 \textwidth]{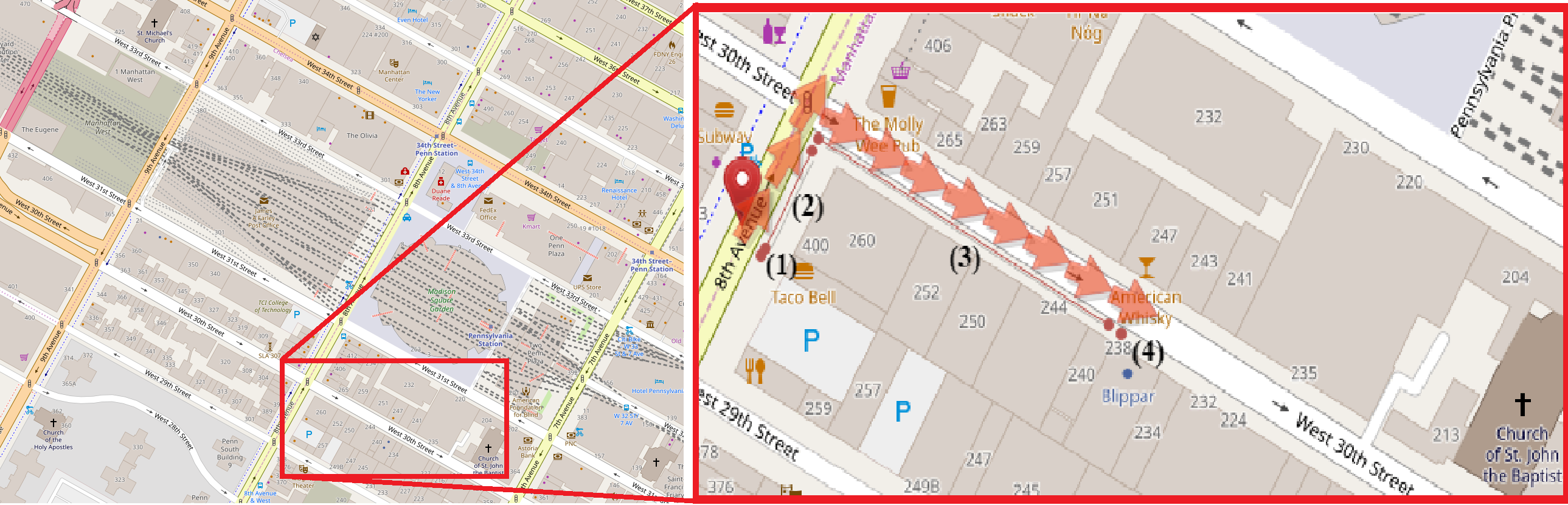}}
            
            \vspace*{-1mm}
                 {\footnotesize{ \begin{spacing}{0.5}
                
                 \begin{flushleft} 
 \begin{mdframed}
 
 \textbf{Instructions:}       
 (1) As you walk out of Taco Bell on 8th Avenue, turn right.
 (2) Then turn right as you reach the intersection of West 30th Street. 
 (3) Now head down West 30th Street for approximately a half block.
 (4) You have gone too far if you reach Church of St. John the Baptist.
 \end{mdframed}
 \end{flushleft} 
\end{spacing} }}

        \caption
        {An Example of a Short Navigation Paragraph: showing a navigation paragraph at the bottom and two images - full map (left image) and a small part of the map (right image). In the full map, many entities are not seen until zoom-in is applied. The navigation paragraph is divided into four sentences: (1) sentence requires a turn action; (2) requires implicit walk actions and an explicit turn; (3) requires walk actions; the last (4) sentence is a verification only and no action is required. 
        } 

        \label{fig:action_example}
    \end{figure*}

We collected the RUN dataset  using
\href{https://www.mturk.com/}{Amazon Mechanical Turk (MTurk)}, allowing  only native English speakers to perform the task. 
 We collected instructions based on three different areas, all in urban, dense parts of Manhattan. The size of each map is $0.5$ $km^{2}$. The dataset contains 2515 navigation instructions (equal to 389 complete paragraphs)  paired with their routes. The  paragraphs are crowd-sourced from 389 different  instructors, of which style and language vary \cite{geva2019we}. 

%
%

Our data collection protocol is as follows. First, we asked the MTurk worker to describe a route between two landmarks of their choice. After having described the complete route in NL, the same worker was instructed to pin their described  route on the map. This  was moderated by showing them the paragraph they narrated, sentence by sentence, so that they have to pin on the map each instruction separately. A worker 
could only pin routes on street paths. 
Furthermore, on every turn the worker had to mark an explicit point on the map which marked the direction in which the worker needs to move next. An example of simple individual instructions and their respective route is given in Figure \ref{fig:action_example}.

We then asked a disjoint set of workers (testers) to verify the routes by displaying the starting point of the route, and displaying the instructions in the paragraph sentence-by-sentence. The tester had to pin the final point of the sentence. Each route was tested by three different workers. Testing the routes allowed us to find incorrect routes (paragraphs that don't match an actual path) and discard them. They also   provide  an estimate of  the human performance on the task (Reported in Section \ref{expirement}, Experiments).  

Having collected the data, we divided the map into   tiles,  each tile is 11.132 m X 11.132 m. Each tile contains the labels of the entities it displays on the map, such as restaurants, traffic-lights, etc., and the walkable streets in it. 
Each walkable street is composed of an ordered list of tiles, including a starting tile and an end tile. 
 Table \ref{tab:linguistic} shows 
statistics over the dataset.
Table \ref{tab:touch} characterize linguistic phenomena in RUN, categorized according to the catalogue of  \citet{chen2018touchdown}. 
 Table \ref{table:statistic_maps} shows a quantitative comparison of the RUN dataset to previous datasets of map-based  navigation.   The table underscores some key features of RUN, relative to the previous tasks.
 RUN contains longer paths and many more entities that are unique, thus appearing for the first time during testing; the size of the map is on a different scale than previous tasks, thus, amplifying the grounding challenge; the number of tiles moved is accordingly larger than in previous datasets, hence increasing the vulnerability to error propagation.


Overall, RUN contains challenging linguistic phenomena, at least as in previous work, and a rich environment, with more realistic paths than in previous tasks.

\section{Models for RUN}
\label{models}

We model RUN as a sequence-to-sequence learning problem, where we map a sequence of instructions to a sequence of actions that should be performed in order to pin the actual path on the map.  
  The execution system we provide for the task defines three types of actions:
\enquote*{TURN}, \enquote*{WALK}, \enquote*{END}.\footnote{RUN has a variation that contains two more types of actions: \enquote*{FACE} is a change of direction to face a specific street and end of the street; \enquote*{VERIFY} gives verification to the current direction when it is explicitly mentioned in the instructions.   For  example,  \enquote{turn right on to 8th Avenue}. However, we have been unsuccessful in benefiting from \enquote{FACE} and \enquote{VERIFY}. We leave them for future research.}
\enquote*{TURN} is one of the following: right-turn, left-turn, turn-around. The turning move is not necessarily an exact 90-degree turn; the execution system looks for the closest turn option.
\enquote*{WALK} is a change of position in the direction we are facing. The streets can be curved, so  \enquote*{WALK}  is relative to the street that the agent is on. Each street is an ordered-list of tiles, so an action of walking two steps is in fact two actions of \enquote*{WALK},  in the direction the agent is facing. The \enquote*{END} action defines the end of each route.
The input consists of an instruction sequence $x_{1:N}$, a map $M$, and a starting point $p_0$ on the map. The output is a sequence of actions $a^*_{1:T}$ to be executed. 
\[
 a^*_{1:T}= arg\max_{a_{1:T}}  P(a_{1:T}|x_{1:N}, M, p_0) \]\[
             = \arg\max_{a_{1:T}} \prod_{t=1}^{T}  P(a_{t}|a_{1:t-1},x_{1:N}, M, p_0) \]
Where \(x_i\) denote sentences, \(a_i\) denotes actions, \(M\) is the map and \(p_0\) is the starting point.


\begin{figure}[t]
\centering
\scalebox{1}{
\scalebox{0.45}{
\includegraphics[width=\textwidth]{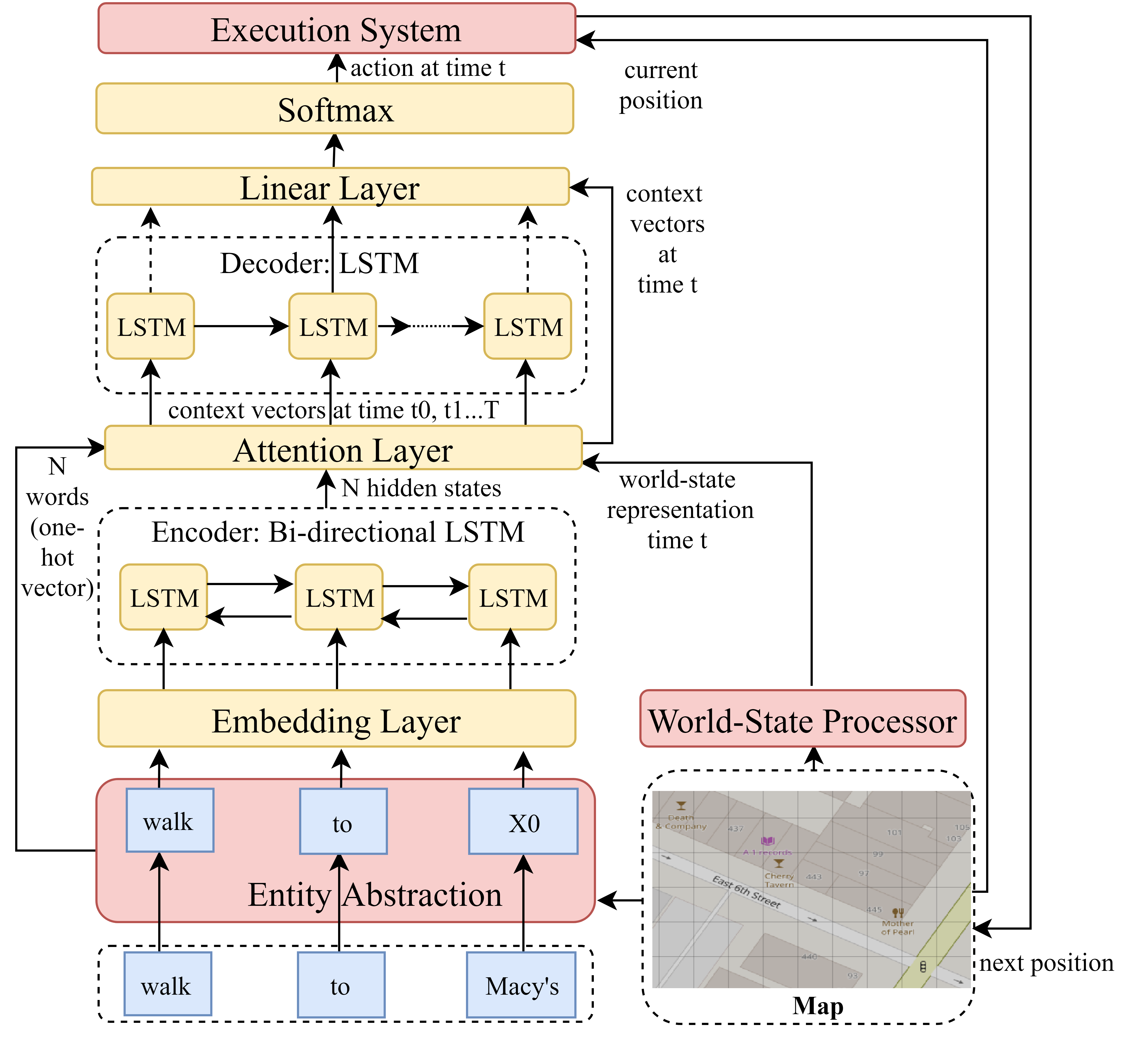}}
}
 \caption{Our Model, Conditioned Generation with Attention over Words and World-States, an Entity Abstraction Layer and an Execution System (CGAEW). The light color (yellow) parts presents a standard Encoder-Decoder with attention, while 
 the dark color (red) are components added on top of a standard CGA.}
\label{fig:model}
\end{figure}

\begin{table*}[th]
\centering
\scalebox{0.8}{
\begin{tabular}{|lll|l|ccc|l|l|}
\cline{1-3} \cline{5-7} \cline{9-9}
\textbf{NO-MOVE}        & \textbf{RANDOM}         & \textbf{JUMP}         &  & \textbf{CGA} & \textbf{CGAE} & \textbf{CGAEW}                    &  & \textbf{HUMAN}          \\ \cline{1-3} \cline{5-7} \cline{9-9} 
30.3\textbackslash 0.3 & 11.2\textbackslash 0.1 & 26.3\textbackslash 0 &  & 43.68 {\small(5.93)}\textbackslash 0.26 & 46.01 {\small(6.13)}\textbackslash 6.17  & 62.37 {\small(3.11)}\textbackslash 10.45 &  & n/a\textbackslash 81.12 \\ \cline{1-3} \cline{5-7} \cline{9-9} 
\end{tabular}

}
\caption{Bounds on Accuracy for Sentences\textbackslash  Paragraphs,  weighted averages over folds (std).}
\label{tab:results}
\end{table*}

\begin{table*}[t]
\scalebox{0.71}{
\begin{tabular}{ll|l|l|l|l}
                       & \textbf{Sentence}                                                                                                  & \textbf{\begin{tabular}[c]{@{}l@{}}JUMP \\ baseline\end{tabular}} & \textbf{CGA} & \textbf{CGAE} & \textbf{CGAEW} \\ \cline{2-6} 
\multicolumn{1}{l|}{1} & \begin{tabular}[c]{@{}l@{}}Just before 9th Avenue, you will see your destination on the right, the West Side Jewish Center.\end{tabular}                       & \cmark                 & \xmark            & \xmark                     & \multicolumn{1}{l|}{\cmark} \\ \cline{2-6} 
\multicolumn{1}{l|}{2} & Turn left onto West 34th Street.                                                                                      & \xmark                 & \cmark            & \cmark                     & \multicolumn{1}{l|}{\cmark} \\ \cline{2-6} 
\multicolumn{1}{l|}{3} & \begin{tabular}[c]{@{}l@{}}At the 8th Avenue and West 20th Street intersection, turn right onto West  20th Street. \end{tabular}                                       & \xmark                 & \xmark             & \cmark                  & \multicolumn{1}{l|}{\cmark} \\ \cline{2-6} 
\multicolumn{1}{l|}{4} & Keep going till you get to the intersection of West 21st Street.                                     & \xmark                 & \xmark            & \xmark                    & \multicolumn{1}{l|}{\cmark} \\ \cline{2-6} 
\multicolumn{1}{l|}{5} & Head west on East 7th for 2 (large) blocks; Its a one-way street.                                                     & \xmark                 & \xmark            & \xmark      & \multicolumn{1}{l|}{\xmark}                            \\ \cline{2-6} 
\end{tabular}
}
\caption{Error analysis of all models, for different instructions, showing succeeded / failure on predicting the path.}
\label{tab:error_analysis}
\vspace{-0.1in}
\end{table*}

Our basic model for  RUN  is a sequence-to-sequence model
similar to the work of \citet{mei2015listen} on   SAIL, and inspired by \citet{xu2015show}.
It is based on Conditioned Generation with Attention (CGA). To this model we added an Entity abstraction layer (CGAE) and a World-state representation (CGAEW). It thus consists of six  components we describe in turn -- Encoder, Decoder, Attention, Entity Abstraction, World-State Processor,  Execution-System. The complete architecture is depicted in Fig.\ \ref{fig:model}.

\textbf{The Encoder}  takes the sequence of words that assembles a single sentence and encodes it as a vector using a biLSTM \citep {graves2005framewise}.  
\textbf{The Decoder} is an LSTM 
generating a sequence of actions that the execution-system can perform, according to weights defined by an Attention layer. 
The \textbf{Entity Abstraction} component deals with  out-of-vocabulary words (OOV). We adopt a similar approach to \citet{iyer2017learning,suhr2018learning},  replacing phrases in the sentences which refer to previously unseen entities 
with variables, prior to delivering the sentence to the Encoder. E.g., \enquote{Walk from Macy's to 7th street} turns into \enquote{Walk from X1 to Y1}.
Variables are typed (streets, restaurants, etc.) and  are numbered based on their order of occurrence in the sentence. {The numbering resets after every utterance, so the model remains with a handful of typed   entity-variables.}  
The \textbf{World-State Processor} 
maps variables to the entities on the map which are mentioned in the sentence. The world-state representation consists of two vectors,  one representing the entities at the current position, and one representing the entities in the path ahead.  
The \textbf{Attention} layer considers the sequence of encoded words {\em as well as} current world-state, and  provides weights on the words for each of the decoder steps. 
In both training and testing, the {\bf Execution-System} executes each action separately to produce the next position.\footnote{Our code, models, complete maps, annotated dataset,  and evaluation: \href{https://github.com/OnlpLab/RUN}{https://github.com/OnlpLab/RUN}.}

\section{Experiments}
\label{expirement}
We  evaluate our model  on  RUN  
and  assess the contribution of the particular components that we added on top of  the standard CGA model.


We train the model using a negative log-likelihood loss, 
and used Adam \citep{kingma2014adam} optimization. For weights initialization we rely on  \citet{glorot2010understanding}. 
We used a grid search to validate the hyper-parameters. The model converged at around 30 epochs and produced good results with 0.9 drop-out and a beam  of size 4. 
During inference, we seek the best candidate path using beam-search and normalize the scores of the sequences according to \citet{wu2016google}.
\par
 We follow the  evaluation methodology defined by  \citet{chen2011learning} for  SAIL  where we use three-fold validation, and in each fold,  we use two maps for training (90\%) and validation (10\%) and test on the third one. We report a sized-weighted average test result. 
%
For all models we report the accuracy per single sentences and full paragraphs. Success is measured by generating an exact route, not striding away from the path. The last position on the path should be  within five tile euclidean distance from the intended destination, as the position explained in the instruction might not be specific enough for one tile.\footnote{We selected this distance as it was the average distance our successful mechanical tester-workers arrived from the intended pinned point.} In single-sentences, the last position should also be facing the correct direction.






We provide three simple baselines for the RUN task:
(1) NO-MOVE: the only position considered is the starting point; 
(2) RANDOM: As in \citet{anderson2018vision}, turn to a randomly selected heading,  then execute a number of \enquote*{WALK} actions of an average route; (3) JUMP: at each sentence, extract entities from the map  
and move between them in the order they appear. If the \enquote*{WALK} action is invalid we take a random \enquote*{TURN} action. 


Table \ref{tab:results}  shows the results for the baseline models as well as the HUMAN measured performance on the task. The human performance provides an upper bound for the RUN task performance, while  the simple baselines provide lower bounds. The best baseline model is  NO-MOVE, reaching an accuracy of 30.3\% on single sentences and 0.3 on complete paragraphs. For the HUMAN case, paragraph accuracy reaches above 80.

Table \ref{tab:results} shows the results of our model as an ablation study, and Table \ref{tab:error_analysis} shows typical errors of each variant. We see that CGAE outperforms CGA, as the swap of entities with variables lowers the complexity of the language that the model needs to learn, allowing the model to effectively cope with unseen entities at test time. We further found that, in many cases, CGAE produces the right type of action, but it does not produce enough of it to reach the intended destination. We attribute these errors to
the absence of a world-state representation, resulting in an incapability to ground instructions to specific locations. CGAEW improves upon CGAE as the existence of world-state in the score of the attention layer allows the model better learn the grounding of entities in the instruction to the map. However 
our best model still fails 
on features not captured by our world-state: abstract unmarked entities  such as blocks, intersections, etc, and generic entities such as traffic-lights (Tab.\ \ref{tab:error_analysis}).


\section{Conclusion}
\label{conclusion}


We introduce RUN, a new task and dataset for NL navigation in realistic urban environments.
We collected 
(and verified) 
NL navigation instructions aligned with actual paths, and propose a strong neural baseline for the task.
Our ablation studies 
  show the significant contribution of each of the components we propose. 
In the future we plan to 
extend the world-state representation, and enable the model to ground  {\em generic} and {\em abstract} concepts as well. We further intend to add additional signals, for instance coming from vision
(cf.\ \citet{chen2018touchdown}), for more accurate localization.
  

\section{Acknowledgments}

We thank Yoav Goldberg and Yoav Artzi, for their advice and comments. We thank the ONLP team at the Open University of Israel, for fruitful discussions throughout the process. We further thank the anonymous reviewers for their helpful comments.
This research was presented at Georgetown University, Cornell University, and Tel-Aviv University.
This research is supported by European Research Council, ERC-StG-2015 scheme, Grant number 677352, and by the Israel Science Foundation (ISF), Grant number 1739/26, for which we are grateful.

\bibliography{acl2019}
\bibliographystyle{acl_natbib}
\end{document}